\documentclass{article}
\usepackage{spconf,amsmath,graphicx,subfigure,booktabs,multirow,epstopdf,setspace}
\usepackage[colorlinks=true, allcolors=blue]{hyperref}

\title{Non-Iterative Simultaneous Rigid Registration Method for Serial Sections of Biological Tissue}
\name{Chang Shu$^{1,2}$, Xi Chen$^{2}$, Qiwei Xie$^{2}$, Chi Xiao$^{1,2}$, Hua Han$^{2,3,4}$\thanks{This paper is supported by Scientific Instrument Developing Project of Chinese Academy of Sciences (No.YZ201671), National Science Foundation of China (NO. 61201050) and Special Program of Beijing Municipal Science and Technology Commission (No.Z161100000216146).}}
\address{$^{1}$ University of Chinese Academy of Sciences, Beijing, China\\
     $^{2}$  Institute of Automation, Chinese Academy of Sciences, Beijing, China\\
     $^{3}$ School of Future Technology, University of Chinese Academy of Sciences, Beijing, China\\
     $^{4}$ The Center for Excellence in Brain Science and Intelligence Technology, CAS, Shanghai, China\\}

\begin{document}
\begin{spacing}{1.0}

\maketitle

\begin{abstract}
In this paper, we propose a novel non-iterative algorithm to simultaneously estimate optimal rigid transformation for serial section images, which is a key component in volume reconstruction of serial sections of biological tissue.
In order to avoid error accumulation and propagation caused by current algorithms, we add extra condition that the position of the first and the last section images should remain unchanged.
This constrained simultaneous registration problem has not been solved before.
Our algorithm method is non-iterative, it can simultaneously compute rigid transformation for a large number of serial section images in a short time.
We prove that our algorithm gets optimal solution under ideal condition.
And we test our algorithm with synthetic data and real data to verify our algorithm's effectiveness.
\end{abstract}
\section{Introduction}
Volume reconstruction from serial sections of biological tissue \cite{briggman2012volume,helmstaedter2013cellular,Micro} has drawn great attention in the community of neuroscience in recent years.
Due to the distortion caused by sectioning, microscopic image registration which aims to recover the 3D continuity of serial sections becomes a key component therein.

Several 3D registration methods \cite{wang2015fully,rossetti2017dynamic,ourselin2001reconstructing,schmitt2007image,pichat2015multi} have been proposed for serial section images.
Since reliable correspondences could only be extracted from adjacent section images, these methods always choose one of section images as reference, and then do forward or backward image registration sequentially for every two neighboring images.
These sequential methods alleviate the difficulty of 3D registration, but introduce error accumulation and propagation.
And they are all time consuming making them inappropriate for simultaneously registering large numbers of section images.

Actually, the correct position of the first and the last section images can be known by taking photos of top and bottom surfaces of sample before sectioning.
Assuming that positions of the first and the last section images have been correctly adjusted, if optimal transformation of remaining section images could be simultaneously estimated, error would not accumulate and propagate.
Therefore it makes sense to achieve simultaneous registration for serial section images under the condition that the position of the first and the last section images are fixed.
This constrained simultaneous registration problem is brand new, it has not been solved before.

To address above problem, we propose a novel non-iterative method to simultaneously estimate optimal rigid transformation for serial section images, while limiting the position of the first and the last section images to remain unchanged.
We prove that our algorithm can get optimal solution when condition is ideal.

\section{non-iterative simultaneous rigid registration method}
First of all, we transform the first and the last images into correct positions according to their relative position before sectioning.
We utilize SIFT flow method \cite{siftflow} to obtain robust correspondences between adjacent images.
SIFT flow algorithm is able to match densely sampled, pixel-wise SIFT features between two images.
Even though individual feature has low discrimination, its correspondence can be inferred by taking into account neighborhood relationship.
From extracted dense correspondences, we further select sparse pairs with low matching error.
In this way, the correspondences we obtain is robust and reliable.
What remains to do is solving a constrained simultaneous registration problem.

Above problem can mathematically described as follows:
Point set ${\textbf{X}_{i,j}}$ is composed of the landmarks extracted from i-th image, whose correspondences is ${\textbf{X}_{j,i}}$ from the j-th image.
Since they are one-to-one correspondence, so the number of ${\textbf{X}_{i,j}}$ equals that of ${\textbf{X}_{j,i}}$, denoted by ${m_i}$.
Our goal turns to be estimating optimal rigid transformations for those point sets to minimize corresponding point distances across them, under the condition that the position of ${\textbf{X}_1}$ and ${\textbf{X}_n}$ are fixed.
So the objective can be defined as
\begin{equation}\label{1}
\begin{array}{*{20}{l}}
{\mathop {\arg \min }\limits_{{{\bf{R}}_i},{{\bf{T}}_i}} \sum\limits_{i = 1}^{n - 1} {{{\left\| {{{\bf{R}}_i}{{\bf{X}}_{i,i + 1}} + {{\bf{T}}_i}{{\bf{e}}_{i}} - {{\bf{R}}_{i + 1}}{{\bf{X}}_{i + 1,i}} - {{\bf{T}}_{i + 1}}{{\bf{e}}_{i}}} \right\|}_F}^2} }\\
{s.t.\left\{ {\begin{array}{*{20}{c}}
{{\bf{R}}_i^T{{\bf{R}}_i} = {\bf{I}}}\\
{\det ({{\bf{R}}_i}) = 1}\\
{{{\bf{R}}_1} = {{\bf{R}}_n} = {\bf{I}}}\\
{{{\bf{T}}_1} = {{\bf{T}}_n} = {{\bf{0}}_{2 \times 1}}}
\end{array}} \right.}
\end{array}
\end{equation}
Where rigid transformation of point set ${\textbf{X}_{i,j}}$ is represented by the combination of rotation matrix ${{\bf{R}}_i}$ and translation vector ${{\bf{T}}_i}$.
${{\bf{0}}_{2 \times 1}}$ is a zero vector of size $2 \times 1$,
and ${{\bf{e}}_{i}}$ is a row vector of length ${m_i}$ with each component equal to one.

In following subsection, we will explain our approach to estimate rotation and translation, provide work flow of our method and discuss under what condition we can get optimal solution.

\subsection{Estimation of Rotation Matrix}
We move the centroid of every point set to the origin, and denote the point set after translation as $\widehat {{X_{i,j}}}$.
This process is referred to as centralization of ${\textbf{X}_{i,j}}$.
We optimize (\ref{2}) to estimate rotation matrix.
It will be proven that under ideal condition, (\ref{2}) is equivalent to (\ref{1}).
\begin{equation}\label{2}
\begin{array}{*{20}{l}}
{\mathop {\arg \min }\limits_{{{\bf{R}}_i}} \sum\limits_{i = 1}^{n - 1} {{{\left\| {{{\bf{R}}_i}\widehat {{{\bf{X}}_{i,i + 1}}} - {{\bf{R}}_{i + 1}}\widehat {{{\bf{X}}_{i + 1,i}}}} \right\|}_F}^2} }\\
{s.t.\left\{ {\begin{array}{*{20}{c}}
{\begin{array}{*{20}{l}}
{{\bf{R}}_i^T{{\bf{R}}_i} = I}\\
{\det ({{\bf{R}}_i}) = 1}
\end{array}}\\
{{{\bf{R}}_1} = {{\bf{R}}_n} = I}
\end{array}} \right.}
\end{array}
\end{equation}
\\
Let ${\textbf{W}_i} = {\textbf{R}_i}^T{\textbf{R}_{i + 1}},{\textbf{A}_i} = \widehat {{\textbf{X}_{i + 1,i}}}{\widehat {{\textbf{X}_{i,i+1}}}^T}$,
so the minimization can be rewritten as follows:
\begin{equation}\label{3}
\begin{array}{*{20}{l}}
{\mathop {\arg \min }\limits_{{{\bf{W}}_i}} \sum\limits_{i = 1}^{n - 1} { - tr\left( {{{\bf{W}}_i}{{\bf{A}}_i}} \right)} }\\
{s.t.\left\{ {\begin{array}{*{20}{l}}
{\begin{array}{*{20}{c}}
{\prod\limits_{i = 1}^n {{{\bf{W}}_i}}  = {\bf{I}}}\\
{{\bf{W}}_{\rm{i}}^T{{\bf{W}}_i} = {\bf{I}}}
\end{array}}\\
{\det ({{\bf{W}}_i}) = 1}
\end{array}} \right.}
\end{array}
\end{equation}
By using \textbf{Lemma 1} in the appendix, solving (\ref{3}) is equivalent to solving (\ref{4}).
\begin{equation}\label{4}
\begin{array}{*{20}{l}}
{\mathop {\arg \min }\limits_{{\theta _i}} \sum\limits_{i = 1}^{n - 1} { - tr\left( {{{\bf{C}}_i}{{\bf{S}}_i}} \right)\cos {\theta _i}} }\\
{s.t.\sum\limits_{i = 1}^{n{\rm{ - }}1} {{\theta _i} =  - \theta } }
\end{array}
\end{equation}

Optimal solution to (\ref{3}) equals ${{\bf{H}}_i}{{\bf{V}}_i}{{\bf{C}}_i}{\bf{U}}_i^T$.
Where ${{\bf{H}}_i}$ is a $2 \times 2$ rotation matrix whose rotation degree is ${\theta _i}$, which is the solution of (\ref{4}).
And (\ref{4}) can be easily solved by using interior point method.
${{\bf{U}}_i}{{\bf{S}}_i}{\bf{V}}_i^T$ is Singular Value Decomposition (SVD) of $\textbf{A}_i$.
${{\bf{C}}_i}$ is a diagonal matrix $diag(1,\det ({{\bf{V}}_i}{\bf{U}}_i^T))$.
$\theta$ is the rotation degree of $\prod\limits_{i = 1}^n {{{\bf{V}}_i}{{\bf{C}}_i}{\bf{U}}_i^T} $.
We provide lemmas and their proofs in the appendix.
\subsection{Estimation of Translation Vector}
With the rotation matrix estimated,
we define ${{\bf{Z}}_i} = {{\bf{R}}_i}{{\bf{X}}_{i,i+1}} - {{\bf{R}}_{i + 1}}{{\bf{X}}_{i + 1,i}}$,
${\bf{T}}{{\bf{h}}_i} = {{\bf{T}}_i} - {{\bf{T}}_{i + 1}}$,
minimization of function in (\ref{1}) can be rewritten as
\begin{equation}\label{10}
\begin{array}{*{20}{l}}
{\mathop {\arg \min }\limits_{{\bf{T}}{{\bf{h}}_i}} \sum\limits_{i = 1}^{n - 1} {{{\left\| {{{\bf{Z}}_i} + {\bf{T}}{{\bf{h}}_i}{{\bf{e}}_i}} \right\|}_F}^2} }\\
{s.t.\sum\limits_{i = 1}^{n - 1} {{\bf{T}}{{\bf{h}}_i}}  = {{\bf{0}}_{2 \times 1}}}
\end{array}
\end{equation}
Equating the partial derivative of (\ref{10}) with respect to $\textbf{Th}_i$ to zero, and using \textbf{Sherman-Morrison Formula} \cite{Hager1989Updating}, we obtain
\begin{equation}\label{13}
{\bf{T}}{{\bf{h}}_i} =  - m_i^{ - 1}{{\bf{Z}}_i}{\bf{e}}_i^T + \frac{{m_i^{ - 1}}}{{m_1^{ - 1} +  \cdots  + m_{n - 1}^{ - 1}}}\sum\limits_{j = 1}^{n - 1} {m_j^{ - 1}{{\bf{Z}}_j}{\bf{e}}_j^T}
\end{equation}
so
\begin{equation}\label{14}
{{\bf{T}}_1} = {{{\bf{0}}_{2 \times 1}}},{{\bf{T}}_i} = \sum\limits_{j = 2}^i {{\bf{T}}{{\bf{h}}_{j - 1}}} ,i = 2, \cdots ,n
\end{equation}
\subsection{Algorithm Flow}
We summarize our non-iterative simultaneous rigid registration algorithm in \textbf{Algorithm 1}.
\\
\noindent\rule{0.48\textwidth}{1.5pt}
\noindent  \textbf{Algorithm 1: }Non-iterative Simultaneous Rigid Registration
\vspace{-0.1in}
\\
\noindent\rule{0.48\textwidth}{1.5pt}
\\
\noindent {\bfseries input:}  original serial section images ${{\bf{I}}_i} \quad (i = 1, \ldots ,n)$
\vspace{-0.1in}
\\
\noindent\rule{0.48\textwidth}{0.5pt}
\\
\textbf{1:}\quad ${\textbf{X}_{i,i+j}}$ $\longleftarrow$ landmarks of ${{\bf{I}}_i}$ \quad ($i = 1, \ldots ,n;j=-1,1$)
\\
\textbf{2:}\quad $\widehat {{\textbf{X}_{i,i+j}}}$ $\longleftarrow$ centralized ${\textbf{X}_{i,i+j}}$ \ ($i = 1, \ldots ,n;j=-1,1$)
\\
\textbf{3:}\quad ${\textbf{A}_i} $ $\longleftarrow$ $ \widehat {{\textbf{X}_{i + 1,i}}}{\widehat {{\textbf{X}_{i,i+1}}}^T}$ \quad ($i = 1, \ldots ,n-1$)
\\
\textbf{4:}\quad $\textbf{V}_i$,$\textbf{S}_i$,$\textbf{U}_i$ $\longleftarrow$ SVD of $\textbf{A}_i$ \quad ($i = 1, \ldots ,n-1$)
\\
\textbf{5:}\quad $\textbf{C}_i$ $\longleftarrow$ $diag(1,\det ({\textbf{V}_i}\textbf{U}_i^T))$ \quad ($i = 1, \ldots ,n-1$)
\\
\textbf{6:}\quad ${\theta_i}$ $\longleftarrow$ solution of (\ref{4}) \quad ($i = 1, \ldots ,n-1$)
\\
\textbf{7:}\quad $\textbf{H}_i$ $\longleftarrow$ $\left[ {\begin{array}{*{20}{c}}
{\cos {\theta _i}}&{ - \sin {\theta _i}}\\
{\sin {\theta _i}}&{\cos {\theta _i}}
\end{array}} \right]$ \quad ($i = 1, \ldots ,n-1$)
\\
\textbf{8:}\quad ${\textbf{W}_i}$ $\longleftarrow$ ${\textbf{H}_i}{\textbf{V}_i}{\textbf{C}_i}\textbf{U}_i^T$ \quad ($i = 1, \ldots ,n-1$)
\\
\textbf{9:}\quad $\textbf{R}_1$ $\longleftarrow$ ${\bf{I}}$, $\textbf{R}_i$ $\longleftarrow$ $\prod\limits_{k = 1}^{i - 1} {{\textbf{W}_k}} $ \quad ($i = 2, \ldots ,n$)
\\
\textbf{10:}\quad ${\textbf{Z}_i}$ $\longleftarrow$ ${\textbf{R}_i}{\textbf{X}_{i,i+1}} - {\textbf{R}_{i + 1}}{\textbf{X}_{i + 1,i}}$ \quad ($i = 1, \ldots ,n-1$)
\\
\textbf{11:}\quad ${\textbf{Th}_i}$ $\longleftarrow$  $ - m_i^{ - 1}{{\bf{Z}}_i}{\bf{e}}_i^T + \frac{{m_i^{ - 1}}}{{m_1^{ - 1} +  \cdots  + m_{n - 1}^{ - 1}}}\sum\limits_{j = 1}^{n - 1} { - m_i^{ - 1}{{\bf{Z}}_i}{\bf{e}}_i^T}$ \\ ($i = 1, \ldots ,n-1$)
\\
\textbf{12:}\quad $\textbf{T}_1$ $\longleftarrow$ ${{{\bf{0}}_{2 \times 1}}}$, ${\textbf{T}_i}$ $\longleftarrow$ $\sum\limits_{j = 2}^i {{\bf{T}}{{\bf{h}}_{j - 1}}}$ \quad ($i = 2, \ldots ,n$)
\\
\textbf{13:}\quad ${\textbf{I}_{i}}$ $\longleftarrow$ ${\textbf{I}_{i}}$ transformed by ${\textbf{R}_{i}}$ and ${\textbf{T}_{i}}$  \quad ($i = 1, \ldots ,n$)
\\
\noindent\rule{0.48\textwidth}{0.5pt}
\\
\noindent{\bfseries output:}  registered serial section images ${{\bf{I}}_i} \quad (i = 1, \ldots ,n)$
\vspace{-0.1in}
\\
\noindent\rule{0.48\textwidth}{1.5pt}
\subsection{Optimality Conditions}
If all the images are under rigid transformation, for ${{{\bf{X}}_{i,i+1}}}$ and ${{{\bf{X}}_{i+1,i}}}$ are one-to-one correspondences, we can opine that they are just the same point set ${{\bf{S}}_{i}}$ under different rigid transformations.
Let ${{\bf{X}}_{i,i+1}} = {{\bf{r}}_i}{{\bf{S}}_{i}} + {{\bf{t}}_i}$ and ${{\bf{X}}_{i+1,i}} = {{\bf{r}}_{i+1}}{{\bf{S}}_{i}} + {{\bf{t}}_{i+1}}$.
If the position of the first and the last image have been correctly adjusted, we have
$\begin{array}{*{20}{l}}
{{{\bf{r}}_1} = {{\bf{r}}_n} = {\bf{I}},{{\bf{t}}_1} = {{\bf{t}}_n} = {{\bf{0}}_{2 \times 1}}}
\end{array}$.
Hence (\ref{2}) can be rewritten as
\begin{equation}\label{15}
\begin{array}{*{20}{l}}
{\mathop {\arg \min }\limits_{{{\bf{R}}_i}} \sum\limits_{i = 1}^{n - 1} {{{\left\| {\left( {{{\bf{R}}_i}{{\bf{r}}_i} - {{\bf{R}}_{i + 1}}{{\bf{r}}_{i + 1}}} \right){{\bf{S}}_i}} \right\|}_F}^2} }\\
{s.t.\left\{ {\begin{array}{*{20}{c}}
{\begin{array}{*{20}{l}}
{{\bf{R}}_i^T{{\bf{R}}_i} = I}\\
{\det ({{\bf{R}}_i}) = 1}
\end{array}}\\
{{{\bf{R}}_1} = {{\bf{R}}_n} = I}
\end{array}} \right.}
\end{array}
\end{equation}
Here we have a trivial solution to (\ref{15}):
\begin{equation}\label{16}
{{\bf{R}}_i} = {\bf{r}}_i^{ - 1}{\rm{ }} \quad (i = 2, \ldots ,n - 1)
\end{equation}
Substitute (\ref{16}) into (\ref{13}), (\ref{13}) is simplified as
\begin{equation}\label{17}
{\bf{T}}{{\bf{h}}_i} =  - m_i^{ - 1}{{\bf{Z}}_i}{\bf{e}}_i^T
\end{equation}
Substitute (\ref{17}) into (\ref{1}), we get (\ref{2}), proving that under ideal condition, it is equivalent to use (\ref{2}) to replace (\ref{1}).

So if we want to get optimal solution, we have to satisfy two conditions: firstly all the serial sections are rigidly deformed, secondly the position of the first and the last image is correctly adjusted, which means that they have no relative displacement.
Our algorithm is not designed to handle nonrigid deformation, but it can still offer robust initial estimate for other nonrigid registration algorithms.
\section{Experimental Evaluation}
In order to verify the effectiveness of the algorithm, we test our algorithm with both synthetic and real data.
With the former we prove that even though correspondences are not accurate, we can still register well.
While the latter demonstrates its application to real scenes.
\subsection{Synthetic Data}
\begin{figure}[!ht]
    \centering
    \includegraphics[width=8.6cm]{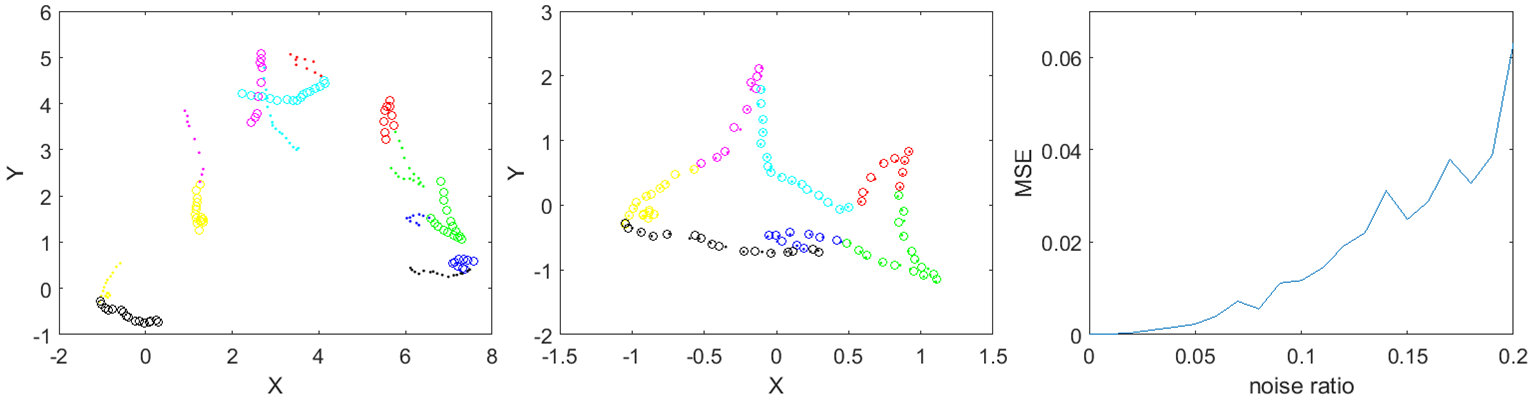}
    \caption{Experimental result. Left: an example of test data. Middle: corresponding registration result. Right: plot of MSE versus noise ratio.}
    \label{fig:1}
\end{figure}

We divide a point set which depicts a fish into 8 subsets, and apply different rigid transformations to them.
Each subset shares partial points with adjacent subset, denoted by the same color.
The first and the last point sets are completely overlapped, so their relative position does not need to be adjusted.
We add random noise proportional to the coordinates of each point to simulate real scenes.
Our experimental result is shown in the Fig.~\ref{fig:1}.
We use mean square error (MSE) between ground truth and our registration result to measure accuracy.
We can see from Fig.~\ref{fig:1} that with noise ratio growing, MSE still maintains at a low level.
Hence we can still get accurate registration result despite getting inaccurate one-to-one correspondences.
\subsection{Real Data}
\begin{figure}[!ht]
    \centering
    \includegraphics[height=2cm]{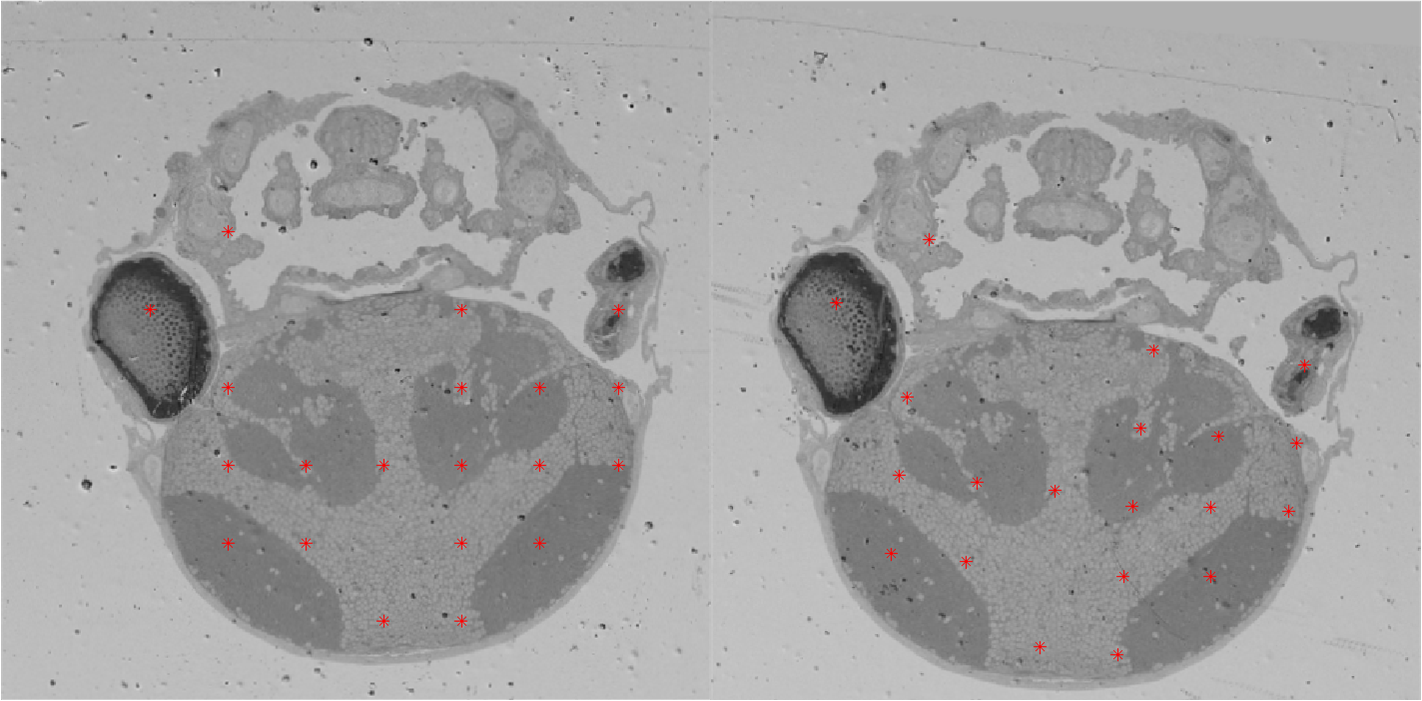}
    \caption{An exhibition of correspondences we extract.
    We utilize SIFT flow method to obtain dense correspondences between adjacent images, then we select sparse pairs with low matching error therein.}
    \label{fig:2}
\end{figure}

We use our non-iterative simultaneous rigid registration algorithm (NSRR) to register serial section images of zebrafish, which contains 336 microscopic images imaged by scanning electron microscope (SEM).
The thickness of the section is 50nm, the resolution of the image is 6144 by 6144 pixels, and the pixel size of the image is 110nm.
Different levels of rotation, translation, scaling and nonlinear deformation are involved.
We utilize SIFT flow method \cite{siftflow} to obtain correspondences illustrated in Fig.~\ref{fig:2}.
\vspace{-0.2in}
\begin{table}[!htbp]
  \centering
  \caption{Comparison of EPE and time}
    \begin{tabular}{|c|c|c|}
    \hline
    Method & EPE & Time \\
    \hline
    3D image registration & 0.0738 & 2184s \\
    \hline
    Pairwise method & 0.0418 & 0.329s \\
    \hline
    NSRR (ours) & 0.0262 & 0.224s \\
    \hline
    \end{tabular}%
  \label{tab:1}%
\end{table}%

\begin{figure}[t]
    \centering
    \includegraphics[width=8cm]{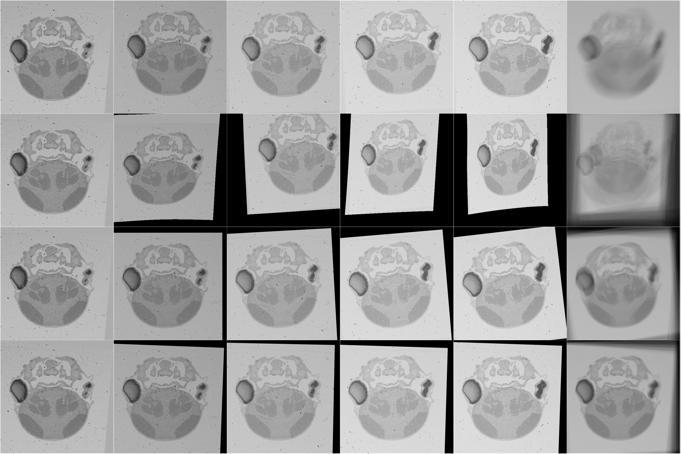}
    \caption{An exhibition of our experimental result. Original data, 3D image registration result, pairwise method result, and our method result are placed separately in the first row to the last row.
    The first five columns show original images and their corresponding registration results, and the last column exhibits average images of the section images before and after registration.}
    \label{fig:3}
\end{figure}

We use endpoint error (EPE) to evaluate accuracy, which is Euclidean distance between adjacent images, averaged over all pixels.
Here an image is resized to a vector.
Pairwise method \cite{arun1987least} and 3D image registration method \cite{wang2015fully} are used as baselines.
Here pairwise method is applied sequentially to register adjacent section images.
Table.~\ref{tab:1} exhibits comparison among 3D image registration method, pairwise method and our method.
Compared with these methods, our method achieves best accuracy with least time.
Fig.~\ref{fig:3} shows their registration results.
We can see from Fig.~\ref{fig:3} that our method achieves best result and does not cause accumulation and propagation of error.
\section{Conclusion}
A constrained simultaneous rigid registration problem of serial section images is presented and solved in this work.
Since reliable correspondences could only be extracted from adjacent section images, current 3D registration algorithms will degenerate into sequentially solving pairwise registration problem, resulting in error accumulation and propagation.

To address this issue, we add a constrain that the first and the last section images must keep their position unchanged, then we estimate optimal rigid transformations for remaining section images to minimize the distance between their correspondences.
The proposed method is non-iterative and can get optimal solution under ideal condition.
It can simultaneously compute rigid transformation for a large number of serial section images in a short time.
\section{Appendix}
{\textbf{Lemma 1:}}
\newline
\emph{
Optimal solution of (\ref{3}) is ${{\bf{W}}_i} = {{\bf{H}}_i}{{\bf{V}}_i}{{\bf{C}}_i}{\bf{U}}_i^T$,
where ${{\bf{U}}_i}{{\bf{S}}_i}{\bf{V}}_i^T$ is the Singular Value Decomposition(SVD) of $\textbf{A}_i$,
${{\bf{C}}_i} = diag(1,\det ({{\bf{V}}_i}{\bf{U}}_i^T))$,
$H_i$ is $2 \times 2$ rotation matrix whose rotation degree ${\theta _i}$ can be obtained by solving (\ref{4}),
and $\theta$ is the rotation degree of $\prod\limits_{i = 1}^n {{{\bf{V}}_i}{{\bf{C}}_i}{\bf{U}}_i^T}$.
}
\vspace{6pt}
\\
{\textbf{The proof of lemma 1:}}
\newline
${{\bf{V}}_i}{{\bf{C}}_i}{\bf{U}}_i^T$ is the optimal rotation matrix in pairwise point sets registration \cite{arun1987least}.
We can assume that ${{\bf{W}}_i} = {{\bf{H}}_i}{{\bf{V}}_i}{{\bf{C}}_i}{\bf{U}}_i^T$,
where ${\textbf{H}_i}$ is an unknown rotation matrix we need to evaluate.
Using commutative law of multiplication for rotation matrix, the constraint in (\ref{3}) can be represented as
\begin{equation}\label{5}
\prod\limits_{i = 1}^n {{{\bf{W}}_i}}  = \prod\limits_{i = 1}^n {{{\bf{V}}_i}{{\bf{C}}_i}{\bf{U}}_i^T{{\bf{H}}_i}}  = \prod\limits_{i = 1}^n {{{\bf{V}}_i}{{\bf{C}}_i}{\bf{U}}_i^T} \prod\limits_{i = 1}^n {{{\bf{H}}_i}}  = {\bf{I}}
\end{equation}
Denoting the rotation degree of $\prod\limits_{i = 1}^n {{{\bf{V}}_i}{{\bf{C}}_i}{\bf{U}}_i^T}$ to be $\theta $ and the rotation degree of ${H_i}$ to be ${\theta _i}$, we obtain $\sum\limits_{i = 1}^n {{\theta _i}}  =  - \theta $.
Using \textbf{Lemma 2}, the objective function in (\ref{3}) becomes:
\begin{equation}\label{6}
\begin{array}{*{20}{l}}
{\sum\limits_{i = 1}^{n - 1} {tr\left( {{{\bf{W}}_i}{{\bf{A}}_i}} \right)}  = \sum\limits_{i = 1}^{n - 1} {tr\left( {{{\bf{C}}_i}{{\bf{S}}_i}{\bf{V}}_i^T{{\bf{H}}_i}{{\bf{V}}_i}} \right)} }\\
{{\quad \quad \quad \quad \quad \quad \, \,} = \sum\limits_{i = 1}^{n - 1} {tr\left( {{{\bf{C}}_i}{{\bf{S}}_i}\left[ {\begin{array}{*{20}{c}}
{\cos  \pm {\theta _i}}&{ - \sin  \pm {\theta _i}}\\
{\sin  \pm {\theta _i}}&{\cos  \pm {\theta _i}}
\end{array}} \right]} \right)} }\\
{{\quad \quad \quad \quad \quad \quad \, \,} = \sum\limits_{i = 1}^{n - 1} { - tr\left( {{{\bf{C}}_i}{{\bf{S}}_i}} \right)\cos {\theta _i}} }
\end{array}
\end{equation}
Therefore solving (\ref{3}) is equivalent to solving (\ref{4}).
\vspace{6pt}
\newline
{\textbf{Lemma 2:}}
\newline
\emph{
With orthogonal matrix \textbf{O} and rotation matrix \textbf{P},
we have ${{\bf{O}}^T}{\bf{PO}} = {\bf{P}}$ or $ {{\bf{P}}^T}$.
}
\vspace{6pt}
\\
{\textbf{proof of lemma 2:}}
\newline
Since \textbf{O} is an orthogonal matrix,
it is either a rotation matrix or a reflection matrix.
If O is a rotation matrix, using commutative law of multiplication for rotation matrix,
${{\bf{O}}^T}{\bf{PO}} = {\bf{P}}{{\bf{O}}^T}{\bf{O}} = {\bf{P}}$.
If \textbf{O} is a reflection matrix,
We denote $\textbf{Rot}(\theta )$ as a rotation matrix with rotation degree $\theta$,
and ${\rm{\textbf{Ref}}}(\phi )$ as a reflection matrix with reflection degree $\phi$,
that is
\begin{equation}\label{7}
\begin{array}{*{20}{l}}
\begin{array}{l}
{\bf{Rot}}(\theta ) = \left[ {\begin{array}{*{20}{c}}
{\cos \theta }&{ - \sin \theta }\\
{\sin \theta }&{\cos \theta }
\end{array}} \right]\\

\end{array}\\
{{\bf{Ref}}(\phi ) = \left[ {\begin{array}{*{20}{c}}
{\cos 2\theta }&{\sin 2\theta }\\
{\sin 2\theta }&{ - \cos 2\theta }
\end{array}} \right]}
\end{array}
\end{equation}
It is known that:
\begin{equation}\label{8}
\left\{ {\begin{array}{*{20}{l}}
{{\bf{Rot}}(\theta ){\bf{Ref}}(\phi ) = {\bf{Ref}}(\phi  + \theta /2)}\\
{{\bf{Ref}}(\theta ){\bf{Ref}}(\phi ) = {\bf{Rot}}(2(\theta  - \phi ))}
\end{array}} \right.
\end{equation}
Therefore
\begin{equation}\label{9}
{{\bf{O}}^T}{\bf{PO}} = {\bf{Ref}}(\phi ){\bf{Rot}}(\theta ){\bf{Ref}}(\phi ) = {\bf{Rot}}( - \theta ) = {{\bf{P}}^T}
\end{equation}

\newpage
\bibliographystyle{IEEEbib}
\bibliography{refs}
\end{spacing}
\end{document}